\documentclass[letterpaper, 10 pt, conference]{ieeeconf}  

\IEEEoverridecommandlockouts                              

\usepackage{amsmath,amssymb}
\usepackage{wasysym}
\usepackage{graphicx}
\usepackage[utf8]{inputenc}
\usepackage[T1]{fontenc}
\usepackage{textcomp}

\title{\LARGE \bf
Modular Robotic Catheters for Endovascular Aneurysm Repair
}

\author{Alex Ranne$^{1*}$, Jinshi Zhao$^{2*}$, Ali Anil Demircali$^{2}$, Songli Moey$^{1}$, Ayhan Aktas$^{3}$,\\ Burak Temelkuran$^{2}$, Nassir Navab$^{4}$, and Ferdinando Rodriguez y Baena$^{1}$
\thanks{$^{*}$ These authors contributed equally to this work.}%
\thanks{Alex Ranne is supported by the UKRI CDT in AI for Healthcare under Grant EP/S023283/1, the ICL-TUM Joint Academy of Doctoral Studies (JADS) program, and by the InnoHK
Initiative of the Innovation and Technology Commission of the Hong Kong
Special Administrative Region Government, Multi-Scale Medical Robotics
Centre, The Chinese University of Hong Kong. Jinshi Zhao is supported by Cancer Research UK under the grant
EDDPJT-May21/100001. }
\thanks{$^{1}$Alex Ranne, Songli Moey, and Ferdinando Rodriguez y Baena are with the Hamlyn Centre for Robotic Surgery, Department of Mechanical Engineering, Imperial College London, SW7 2AZ, UK (Corresponding author: Ferdinando Rodriguez y Baena, e-mail: {\tt\footnotesize f.rodriguez@imperial.ac.uk)}}%
\thanks{$^{2}$Jinshi Zhao, Ali Anil Demircali, and Burak Temelkuran are with the Department of Metabolism, Digestion and Reproduction, Imperial College London, SW7 2AZ, UK.}%
\thanks{$^{3}$Ayhan Aktas is with Department of Mechanical Engineering, Bursa Uludag University,  16059, Bursa, Turkey}
\thanks{$^{4}$Nassir Navab is with the Chair for Computer Aided Medical Procedures and Augmented Reality (CAMP), Technical University of Munich, 85748 Garching, Germany }}%

\begin{document}

\maketitle
\thispagestyle{empty}
\pagestyle{empty}

\begin{abstract}

Fenestrated/Branched endovascular aneurysm repair (FEVAR/BEVAR) require surgeons to navigate catheters and guidewires into various branches of the abdominal aorta, before deploying stent grafts to alleviate pressure on the aneurysm. Previous clinical studies suggests that surgeons continue to struggle with vessel access using standard commercial instruments, prolonging the procedural time and inducing further complications. In this work, we present two contributions to solving this problem: 1) A bespoke 2-segment steerable catheter, consisting of 4 degrees of freedom to enhance dexterity. 2) An expandable, modular tendon-driven actuation platform that can accommodate for the redundancies introduced in our system. To fabricate the catheter, we capitalized on thermal fiber drawing, a technique that creates high-aspect ratio devices at scale, and processed the catheter with laser micro-machining to soften its tip. We evaluated the system using simulations, where we investigated the catheter's bending stiffness, then its steerability with in-vitro experiments in vascular phantoms. This handheld, robotic steerable catheter system has the potential to shorten the length of future endovascular surgeries, and give clinicians the tools to resolve challenging clinical cases.

\end{abstract}

\section{INTRODUCTION}


Cardiovascular diseases (CVD) poses a substantial burden on the world. In 2022, an estimated 19.8 million people passed away due to a case of CVD, constituting 31.59 \% of all deaths worldwide according to the WHO \cite{ihme2023global}. Minimally invasive endovascular surgery (MIES) has become the primary method for treating CVD. In MIES, a small access point is created on the patient's groin, where thin and long medical devices (guidewires and catheters) are introduced via the femoral artery. These devices assist vascular surgeons in navigating through the vasculature with the help of fluoroscopic imaging, then provide an avenue for deploying stent grafts, balloons, or other forms of interventional devices. 

A particular type of CVD, thoraco-abdominal aortic aneurysm (T/AAA) is prevalent in adults over 60 years of age, and has a high mortality rate if ruptured \cite{ullery2018epidemiology}. In the case of T/AAA, the aneurysm sits in the upper part of the aorta, which consists of bifurcations including the renal (to kidney) and mesenteric (to intestines) arteries. As such, surgeons treats the aneurysm with either fenestrated or branched endovascular aneurysm repair (FEVAR/BEVAR), while maintaining blood flow to vital organs. In both cases, the procedure involves the deployment of a main stent graft, a Ni-Ti alloy mesh tube with fabric scaffolding tube and holes cut out on its body, designed to fit the patient's anatomy. This is followed by the deployment of either fenestrations, which are short grafts that guide blood flow into a bifurcating vessel, or long branches that extent fully into each bifurcation to ensure proper blood flow. The choice of FEVAR or BEVAR depends on the extent of the aneurysm, whether it is localized (FEVAR), or widely distributed (BEVAR).

\begin{figure}
    \centering
    \includegraphics[width=0.46\textwidth]{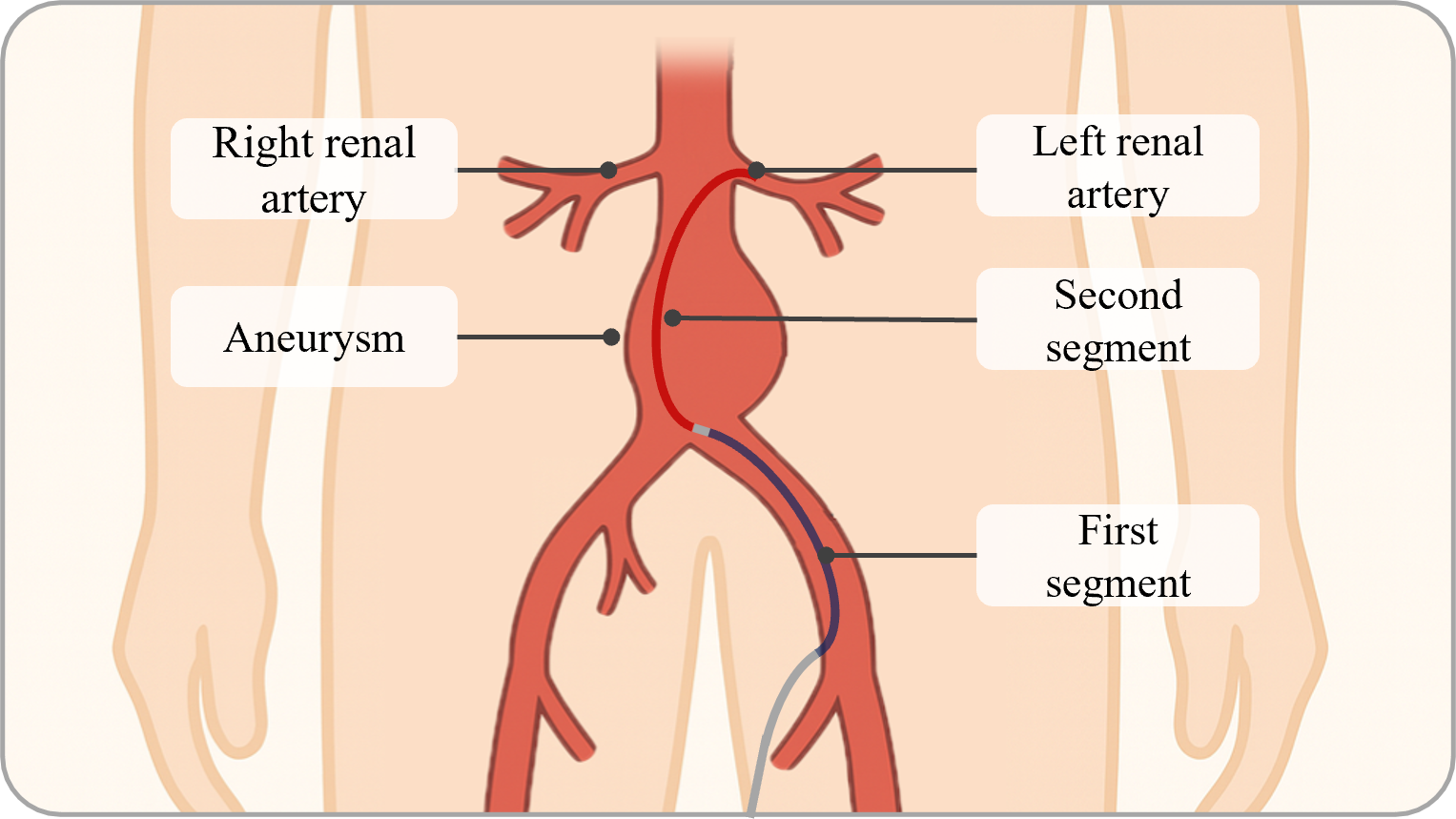}
    \caption{Conceptual rendering of the proposed 2-segment tendon-driven catheter in AAA surgery. }
    \label{fig:Thermal Drawing}
\end{figure}

The challenge of FEVAR/BEVAR lies in target vessel catheterization. Due to vessel tortuosity, unfavorable access angles, and a lack of suitable instruments, these procedures can take an alarmingly long duration, exposing clinicians to high dosage of radiation. A brief FEVAR study in 2021 by Tschischka et al. reported a challenging catheterization of the right renal artery, failing even under multiple access points (femoral and transaxillary) \cite{tschischka2021completion}. The team was only able to access the right renal artery using a steerable sheath, but still faced difficulties in aligning the fenestration with the renal artery, leading to endoleaks in their first deployment attempt. Several multi-center studies on FEVAR/BEVAR reported that in cases of challenging FEVAR/BEVAR cases, the rate of success drops, and subsequent redo of such procedures are necessary \cite{schanzer2020results, karelis2021editor}. In most cases, system failure is associated with difficulty in reaching a target vessel, with a significantly lower success rate of 83\%, compared to the standard statistic of over 95\% \cite{schanzer2020results}. These numbers highlights a clear demand for smaller, advanced and maneuverable instruments that can traverse complex vessel geometries.

To meet these demands, steerable and/or robotic endovascular devices have been developed. These systems have a stiff main body, made from extruded or molded polymers, reinforced with braiding, and paired with a soft tip that can be deflected using a lever on the catheter handle or a robotic device. Commercially, several examples of steerable catheters of size ranging from 2.4 Fr to 12 Fr are applied primarily for cardiac surgery. 
Larger catheter sheath systems, such as the Destino Steerable Sheath by Baylis Medical Techologies \cite{Urbanski2022}, use a braided body to enhance the device's pushability, torquability and kink resistance. The sheath allows for standard catheters and guidewires to be advanced within it. The FlexCath Contour by Medtronic Inc., is designed to perform cryogenic ablation of the atria, without the need to exchange instruments, as the catheter has no working channels \cite{Gammie2010}. However, abandoning a working channel in the catheter sacrifices versatility, therapeutic capability, and the option for contrast injection in exchange for maneuverability. 

Recent research has led to the development of smaller catheters to meet increasingly challenging clinical demands. For example, Abdelaziz et al. introduced a 6 Fr, Magnetic Resonance Imaging (MRI) visible selective steerable catheter, consisting of helically shaped tendon channels, a laser profiled tip to reduce its stiffness and enhance steering capabilities, and a braided body to reduce the chance of kinking. The system was successfully tested in-vivo, providing good coverage in the porcine heart with a controlled backlash of approximately 5 mm \cite{abdelaziz2024thermally}. Commercially, the 2.4 Fr SwiftNINJA system from Merit Medical enables articulation in extreme angles, with multidimensional steering in two planes. This allows for vessel selection without the need for a probing guidewire, which is introduced in standard clinical practice to prevent forceful and aggressive handling of instruments. This system was recently applied to treat congenital heart defects, and selective catheterization of challenging vasculatures in children \cite{haddad2023swiftninja}. Another example uses the 2.5 Fr Benfit Steerable Microcatheter for treating intracranial aneurysms, and successfully deployed neurovascular mesh to fill the aneurysmic cavity in eight separate trials, following navigation in tortuous cerebral arteries \cite{killer2023clinical}. Making the catheter smaller, while retaining its working channel is critical to ensuring sufficient reach. 

In contrast, progress in the development of robotic catheter systems are more limited. In 2012, the Magellan system by Hansen Medical (Mountain view, California, USA) received its FDA approval, and subsequently demonstrated exceptional precision in peripheral artery interventions \cite{rueda2018robotics}. The tendon-driven system excelled in rapid cannulation, using a 7-DOF delta robot as its haptic device to allow for intuitive control of catheter tip. Robotic tendon driven systems can reduce the negative impacts of wire slack on the system performance, and maintain catheter tip positions with its actuators. However, the system fell short of its expectations due to doubts over its cost-effectiveness, large catheter size, and longer setup times \cite{cruddas2021robotic}. Other commercial systems, such as the Niobe Magnetic Navigation system from Steroaxis Inc. (St. Louis, Missouri, USA) consists of similar challenges \cite{condino2022bioengineering}. Making robotic catheters smaller and nimble, and is key to wider adoption of such technology.

In this work, we present a dexterous tendon-driven robotic catheter system. We divide our system development into three phases: scalable fabrication of catheters using thermal fiber drawing, design of a rapid prototyped, modular actuation handle that can accommodate a high number of tendons, and the development of a teleoperation system for device control. Quantitative evaluation is carried out in the form of a finite element method (FEM) simulation to understand the device's bending stiffness, and characterization of its bending angles. To highlight the advantage of a multi-segment system, we then carry out two task-oriented experiments, with one evaluating how well the catheter can traverse a ring obstacle course, and another being an in-vitro study with an abdominal aortic phantom.



\section{SYSTEM DEVELOPMENT} 

\begin{figure}
    \centering
    \includegraphics[width=0.48\textwidth]{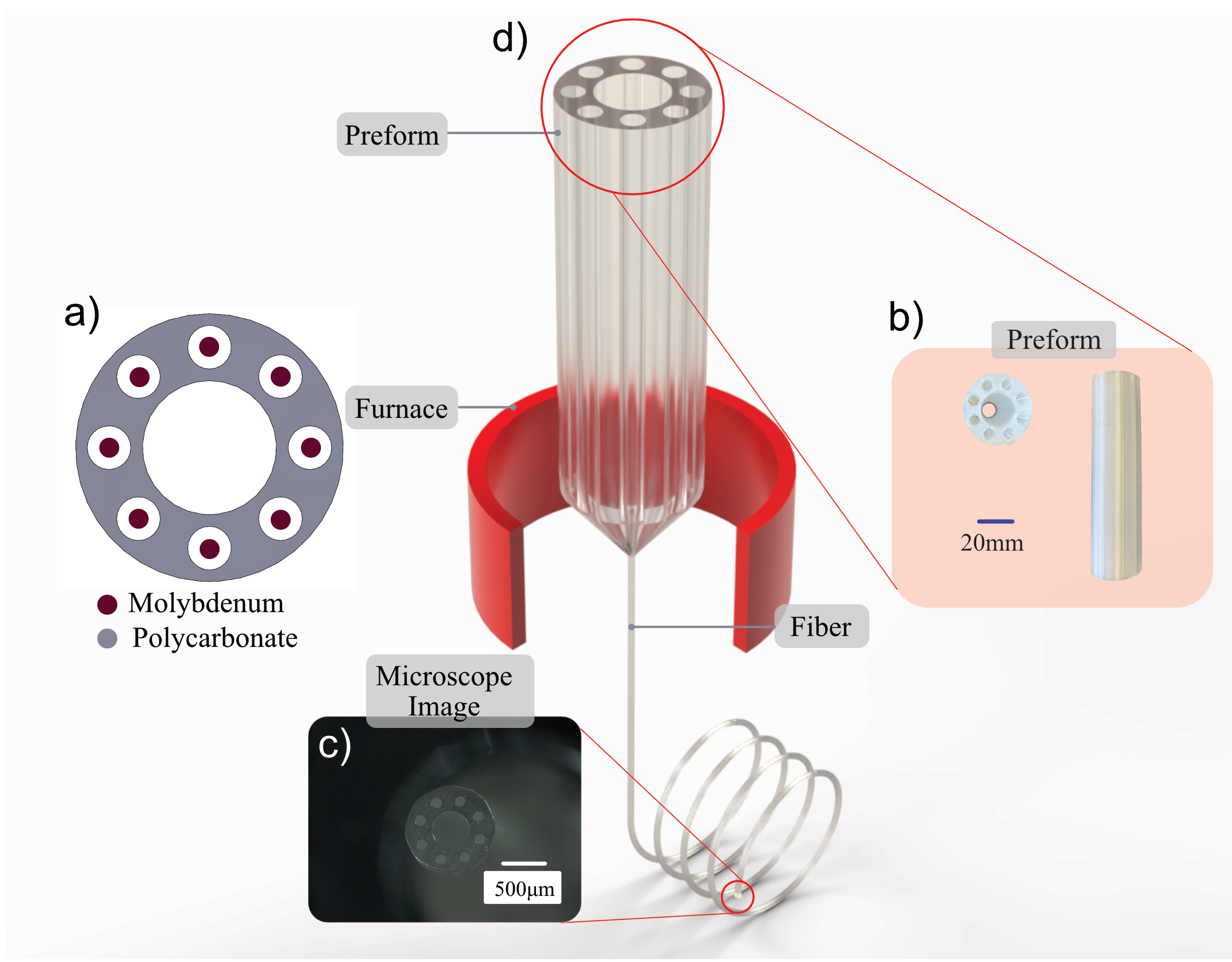}
    \caption{Catheter fabrication a) Modeled catheter design b) Photograph of the printed preform c) Microscope image of the thermally drawn catheter. d) Schematic of the thermal drawing process, where the furnace and the relative position of the preform are shown. The capstan is omitted for clarity. }
    \label{fig:Thermal Drawing}
\end{figure}

\subsection{Scalable fabrication of catheters} \label{Section: Fibre drawing}


Our robotic catheter is fabricated using thermal drawing technology. This technology begins with a preform: a macroscopic bulk material consisting of the same design as the desired fiber, usually made from an amorphous polymer using 3D printing \cite{demircali2025fabrication}. During thermal drawing, the preform is placed in a vertical cylindrical shape furnace, where a section of the preform is heated above the polymer's glass transition temperature $T_g$ to reach its visco-elastic state. Thereafter, tension is applied to the bottom end of the preform, to cause it to neck down, and finally reach the desired cross-sectional size. Fiber drawing has the advantage of retaining intricate preform design without losing structural integrity, while generating tens of meters of fiber in a scalable manner. Compared to extrusion, fiber drawing offers flexibility in creating complex designs with affordable running costs, making it more suitable for prototyping and research phase hardware development.

In this work, we selected polycarbonate (PC, $T_g$ = 112$^oC$, $\alpha$ = $2.1 \times 10^{-5\: o}  C^{-1}$) as the preform material, owing to its sound thermal stability, mechanical strength, dimensional stability, and material availability for 3D printing.

\begin{figure}[h!]
    \centering
    \includegraphics[width=0.5\textwidth]{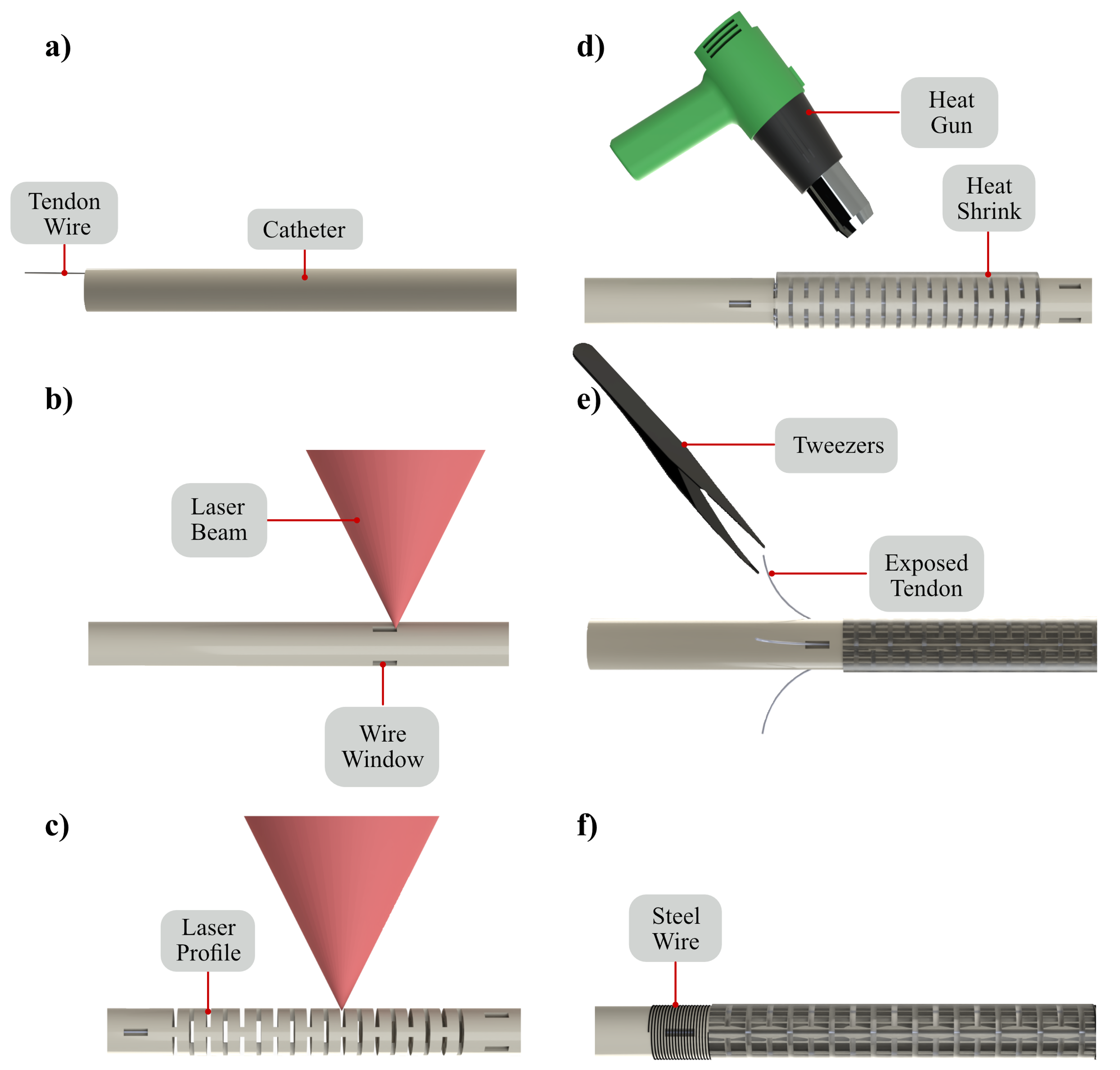}

    \caption{Detailed schematic showing the assembly process for the multi-segment tendon driven catheter. a) Inserting molybdenum wires into the catheter's wire channel b) Laser machining wire windows to expose the tendons for step e). c) Laser machining profiles to reduce the catheters bending stiffness. d) Protecting the laser profile with heat-shrink tubing. e) Expose tendons from lasered windows using a pair of tweezers. f) Securing the tendon's end by sandwiching the wire in between two layers of wounded steel wires of diameter \diameter 50 $\mu m$  }
    \label{fig:Assembly}
\end{figure}


The design of our 6 Fr (2 mm) fiber is presented in Fig. 2(a). It consists of 8 $\times$ 200 $\mu m$ equally distributed tendon channels and a 1.1 mm central guidewire channel, which can accommodate a soft tip guidewire for navigational purposes. The preform has dimensions of 40 mm in diameter and 15 cm in length.
 It is modeled using computer aided design software Solidworks (Dassault Systemes, France), and then 3D printed using a commercial Fused Deposition Modelling (FDM) printer, Ultimaker 3+ Extended (Ultimaker BV, Netherlands). Printing is performed with a 0.4AA print core size, 100\% infill density, triangular infill pattern, and a layer thickness of 0.05 mm. After printing, the printed preform is cured in a 70$^oC$ vacuum oven for 48 hours to remove any residual internal moisture or dissolved gasses. 

During the draw, we define the temperature of a 3-zone furnace to be 120 $^oC$, 190$^oC$ and 85$^oC$, respectively. While the top and middle zone temperatures are above the glass transition temperature, they are empirically chosen to strike a balance between preform viscosity, and structural integrity. The preform downfeed speed is set to 1 mm/min, and the linear speed of the fiber-pulling capstan is set to 0.4 mm/s, but it is fine tuned during the draw based on the measured fiber diameter. Using the conservation of mass, we can determine the fiber diameter $r_d$ using the following equation:

\begin{equation}
    r_d = r_p \sqrt{\frac{v_d}{v_c}}
\end{equation}

Where $r_p$ is the preform diameter, $v_d$ is the preform downfeed rate, and $v_c$ is the capstan rotational speed.

\subsection{Multi-segment steerable catheters}

To provide surgeons with enhanced maneuverability within tortuous vasculature, we propose a one meter long, two segment tendon driven robotic catheter system, using the PC catheter body created from Sect. \ref{Section: Fibre drawing}. 
Prior studies on multi-segmented steerable catheters demonstrated that a two segment system is already suited to create several common catheter shapes \cite{clogenson2015multi}. While there is no study on catheters with an even higher degree of freedom, there also appears to be no clear clinical benefits to such design decision, hence the two segment design was chosen. 

The assembly process for the tendon driven catheter can be summarized in the following steps.

\subsubsection{Tendon Insertion}

Each segment is controlled via four tendons, where antagonistic pairs of tendons (tendons on opposing sides of the catheter) are coupled to the same actuator (see Sect.\ref{Section: Modular Robotic Actuators}) in order to minimize the number of actuators needed, thus lower the cost of the system. We hand-feed 80 $\mu m$ diameter molybdenum alloy wires into each tendon channel after the draw took place (Fig. \ref{fig:Assembly}a). Molybdenum wires are chosen due to its high strength and bio-compatibility, demonstrated by recently studies on Molybdenum based medical devices (eg.\cite{mayers2024insights, toschka2023does}).

\subsubsection{Laser Micromachining}

Tendon-driven mechanisms require rigid coupling between the tendon to the continuum robot. While this is possible using less stiff wires, Molybdenum wires' high strength makes tying knots a challenge. 

To fasten the tendons to their constituent segment, we opted to create small windows on the catheter surface, exposing the wires of interest. Since the wires, and space on the catheter surface available to create the window are at the sub-millimeter level, conventional computer numerically controlled (CNC) machining can no longer meet the tolerances required. As such, a laser micro-machining system (Optec Laser Group,  Belgium, Seraing, Belgium) is adopted. The system has a precision below 3 $\mu m$, capable of controlled abrasion of materials with femtosecond laser pulses. The laser system consists of a microscope and a laser lens, used for visualizing, and performing the cut, respectively. The devices are mounted on a teleoperated x-y stage (with a precision of $\pm$ 10 $\mu m$), suitable for planar sample positioning. To expand this workspace such that a cut can be made along the catheter surface, we built a custom catheter manipulation stage that is capable of translation along its body (linear translation resolution: 0.44 mm) and rotation about its central axis (rotary resolution: 0.0879 $^\circ$). To utilize 4 tendons per segment, we create windows of size 0.2 mm by 0.8 mm at the end of each segment, each being $90^\circ$ apart from each other (Fig. \ref{fig:Assembly}b). 

Along each steerable segment, we laser-profiled the catheter surface with an offset slot design (Fig. \ref{fig:Assembly}c), similar to that found in Abdelaziz et al.'s work \cite{abdelaziz2024thermally}. Laser profiling reduces the force needed to pull the tendons and thus bend the segment, allowing the catheter to bend to tighter angles and improve its maneuverability. Furthermore, to steer into challenging bifurcations and turns, we introduce a difference in bending stiffness between the distal and proximal segment, by varying the cutting depth of each segment. The distal segment is designed to be softer, making it easier to bend to tighter angles, and more responsive. While the proximal segment's purpose is for general positioning that does not require high precision. In this design, we laser machined four sets of profiles, each set being 90 degrees from one another, with a 1mm offset between adjacent profiles, and a 2mm spacing between each slot. The distal segment has a profile depth of 400 $\mu m$, while the proximal segment has a depth of 200 $\mu m$. The choice of these profile designs is justified in Sect. \ref{Section: FEA}.

\subsubsection{Assembly}
The catheter is covered by a thin layer of medical grade heat shrink (wall thickness: 127$\mu m$, ID: 2mm) to protect the moving wires, and prevent any sharp edges on the catheter from damaging the patient's tunica intima, the inner most layer of the blood vessel (Fig. \ref{fig:Assembly}d). 

\begin{figure*}[t!]
    \centering
    \includegraphics[width=0.7\textwidth]{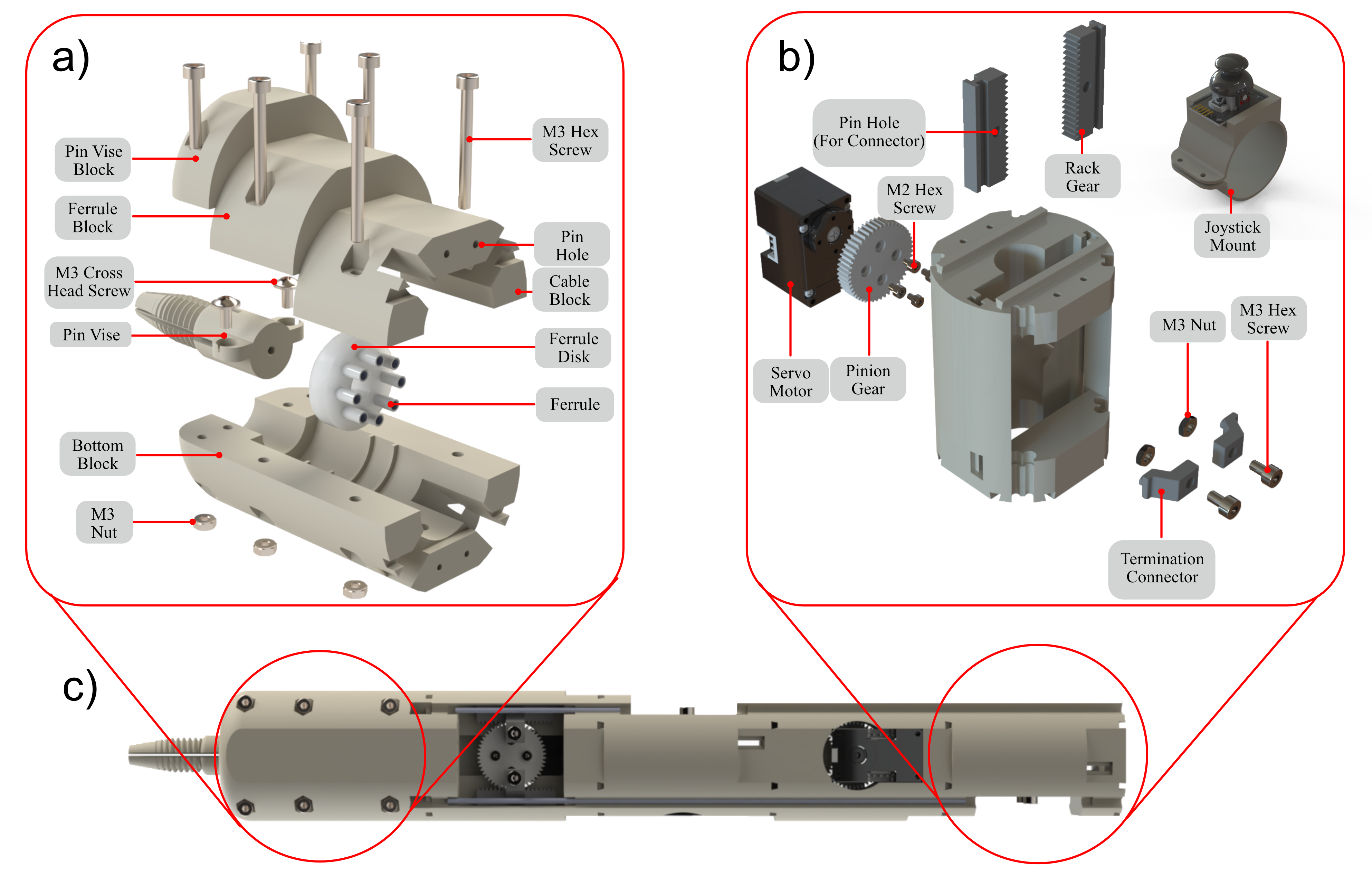}
    \caption{Rendered image of the device handle a) Exploded view of the cap unit, designed to guide the tendons from the catheter into 8 Bowden cables, then into the actuation units. b) Exploded view of the actuation unit, where each unit actuates a pair of tendons on opposite sides of the catheter. The optional joystick module is included in the top right corner. c) Assembled handle, consisting of 4 actuation units and 1 cap unit.}
    \label{fig:Handle}
\end{figure*}

Since we are cutting deep grooves into the catheter, beyond the depth of the wire channel (100 $\mu m$), we pulled the tendons back during the machining process to avoid cutting into them. They are fed back into the channel afterwards, pulled out from the windows using tweezers (Fig. \ref{fig:Assembly} e),  pruned to a short length of approximately 5 mm, then sandwiched and glued between two layers of wounded $50 \mu m$ stainless steel wires secure them to the catheter (Fig. \ref{fig:Assembly} f). Finally, we fastened a pin vise at the end of the catheter for attachment purposes (see Fig. \ref{fig:Handle}).

\subsection{Modular robotic handle} \label{Section: Modular Robotic Actuators}

The robotic catheter handle, as shown in Fig. \ref{fig:Handle}, introduces an expandable platform that is designed to accommodate a high number of tendons. All components of the handle, with the exception of the fasteners and the ferrules, are made from Polylactic acid (PLA) 3D printing, using a Bambu Lab A1 printer (Shenzhen, China). The design proposed in this paper revolves around 3 main modules, the cap module, the actuation module, and the joystick module.

The cap module's task is three-fold: 1) fasten the catheter on the handle, 2) distribute and separate the tendons into their corresponding channels. 3) protect the tendons, while providing the user with the ability to adjust, repair, and reconfigure the tendon configuration. As shown in Fig. \ref{fig:Handle} a), its main body is split into four pieces, a bottom block, which accommodates the catheter/pin vise assembly, and three top pieces, allowing the user to open up any part of the module to access the pin vise (pin vise block), the ferrule disk (ferrule block), and the tendons exiting the cap into the actuators (cable block). The ferrule disk provides a mounting point for spring cables (terminated on both ends with ferrules), which encapsulates the tendons in order to reduce friction with other components, and protect the user from accidentally touching the moving parts. 

When considering the actuation unit, we have decided to shift away from bulky, expensive, and non-intuitive actuation units, to a modular and compact system that can be expanded to accommodate for 2 $\times$ n tendons, where n represents the number of actuators embedded. 

In Fig. \ref{fig:Handle} b), we present the exploded view of an actuation unit. The shape of the module is designed to be mostly cylindrical (\diameter 56 mm), such that when assembled, it can be easily gripped and inserted using one hand. The unit is powered by a Dynamixel XL-330-M288-T, coupled to a 0.5 MOD pinion gear, which is used to drive two 0.5 MOD rack gears on either side with a travel distance of 6cm, converting rotation into translation. Since both racks are in mesh with the same pinion gear, the tendons act in antagonistic pairs, such that when one is pulled, the other slackens. To anchor the end of the tendon to each rack, we added an M3 screw and nut holder, which acts as a clamp when tightened.

To attach individual robotic modules together, we implemented a dovetail sliding joint, consisting of protruding dovetails on the top of the module, and slots on its bottom. While only two rails are needed to accommodate the dovetails, we implemented two sets of rails, with them being perpendicular to each other. This allows us to attach the module in four different orientations, allowing each module to actuate different pairs of tendons. 

One additional feature that is implemented are 4 M-shaped grooves on the module's body, located on four edges of the actuation module. Due to how the handle is designed, tendons that terminate at proximal actuation modules will cross paths with counterparts that terminate at distal modules, implying that there will be two sets of tendons running at each edge at a time. To protect and separate them, the M-shaped grooves provide distinct channels for each tendon, while a steel tube is fitted over the spring cable to prevent the user from accidentally pinching the tendon.

From an ergonomics perspective, surgeons should be able to manually insert and then steer the catheter using only one hand. Therefore, instead of integrating a separate user interface (UI) to control the actuators, such as a PC keyboard or an X-box controller, we instead designed a joystick module that can be fitted onto the handle post assembly. The module consists of four mounting holes for a joystick module, and a partial ring like body, where it can be clamped onto the handle using M3 screws and nuts, which is similar in design to an R-clamp. 

\subsection{Forward Kinematic Model}
The two-segment tendon-driven system can be modeled based on a constant curvature assumption. We assume that, with the help of laser profiling, each bending segment bends uniformly along its length. We also assume that there is no torsional deformation, negligible effect of gravity on the catheter, no external loading and that tendons follow a continuous, parallel path to the backbone.

Since the two segments are offset 45 degrees to each other about the z-axis of the first segment, their coordinate systems can be represented by the following homogeneous transform:

\begin{equation}
    T_{2}^{0} = T_{1}^{0} \cdot R_{z}(45^{\circ})\cdot T_{2}^{1}
\end{equation}

Where $T_{1}^{0}$, $T_{2}^{1}$, and $T_{2}^{0}$  represents the transformation from the base to the tip of the first segment, from the tip of the first segment to the second segment, and from the base to the tip of the second segment, respectively. $R_z(45^\circ)$ is the rotation offset between segments.
We define the homogeneous transformation matrix (HTM) of each segment using the curvature $\kappa$, the arc length $\ell$ and bending plane angle $\phi$:

\begin{equation}
T_{1}^{0} =\begin{bmatrix}
 cos(\phi_1)& -sin(\phi_1)  & 0 & \alpha_1 cos(\phi_1) \\
 sin(\phi_1)& cos(\phi_1) & 0 & \alpha_1 sin(\phi_1) \\
 0 & 0 & 1 & \beta_1 \\
 0 & 0 & 0 & 1 \\
\end{bmatrix}
\end{equation}

\begin{equation}
T_{2}^{1} =\begin{bmatrix}
 cos(\phi_2)& -sin(\phi_2)  & 0 & \alpha_2 cos(\phi_2) \\
 sin(\phi_2)& cos(\phi_2) & 0 & \alpha_2 sin(\phi_2) \\
 0 & 0 & 1 & \beta_2 \\
 0 & 0 & 0 & 1 \\
\end{bmatrix}
\end{equation}

With $\alpha_i = \frac{1-cos(\kappa_i \ell_i)}{\kappa_i}$ and $\beta_i =\frac{sin(\kappa_i \ell_i)}{\kappa_i} $ computed for the $i^{th}$ segment.

\section{EXPERIMENTAL EVALUATION}

\subsection{Finite Element Analysis} \label{Section: FEA}
To better understand the effect of different laser-profiled designs on bending stiffness, we performed mechanical characterization via finite element simulations in COMSOL Multiphysics (Stockholm, Sweden). Using beam theory, the bending stiffness $EI$ of a cantilever beam can be modeled using the following equation:

\begin{equation}
\label{Eq:Bending stiffness}
    EI = \frac{WL^3}{3\delta}
\end{equation}

Where $E$ is the Young's modulus, $I$ is the second moment of area, $W$ is the cantilever end load, $L$ is the length of the beam, and $\delta$ is the end deflection. The addition of laser profiling reduces the bending stiffness substantially by removing material from the catheter's cross section, effectively reducing $I$. 

\begin{figure}
    \centering
    \includegraphics[width=0.46\textwidth]{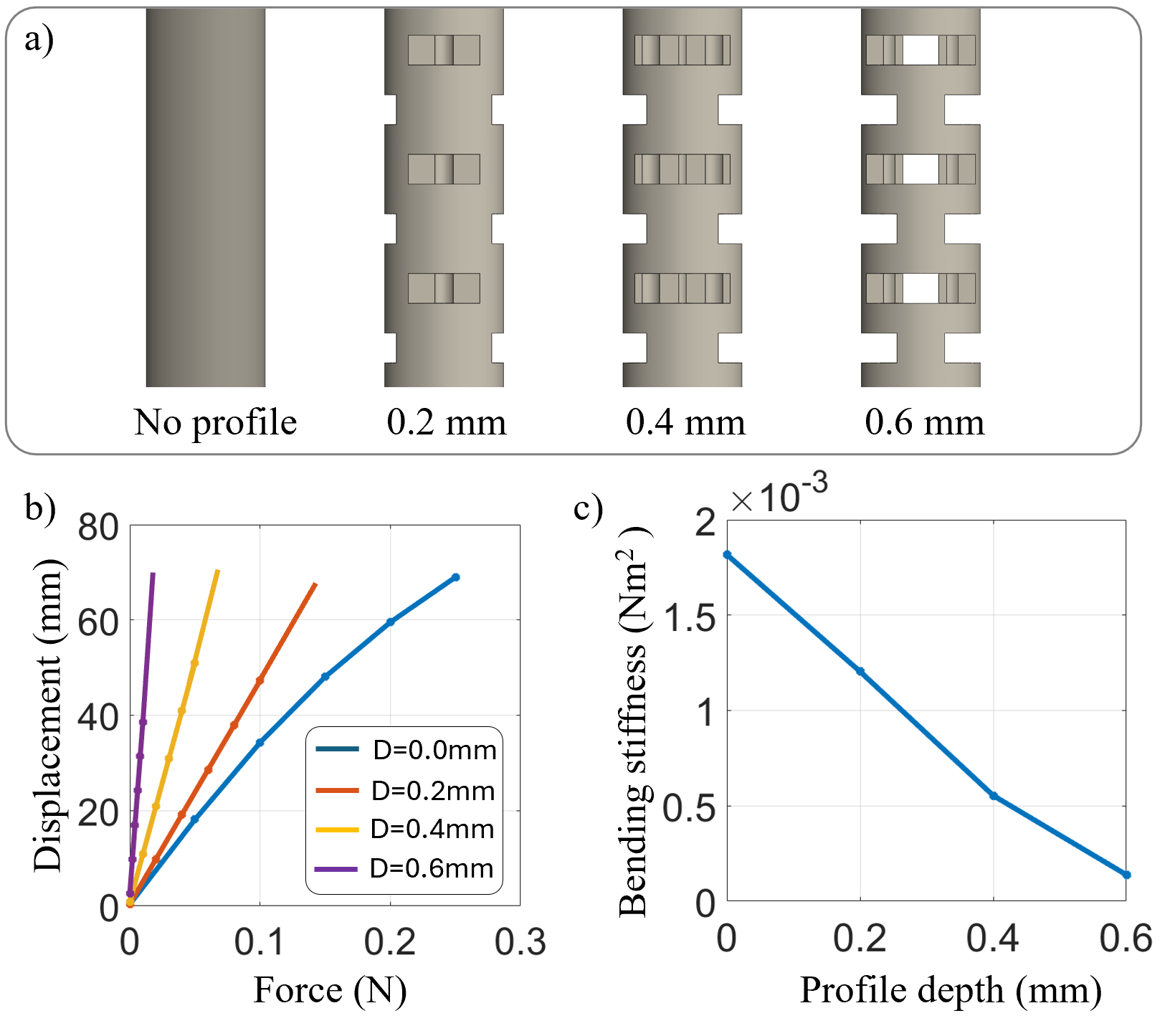}
    \caption{FEA simulation of catheter stiffness for laser profiling optimization. a) Schematic of catheters with varying laser profiling depths. b) Simulation results showing tip displacement under perpendicular force for different laser profiles. c) Calculated bending stiffness of catheters with different laser profiles from the simulation.}
    \label{fig:FEA1}
\end{figure}

To demonstrate this hypothesis, we modeled the steerable tip in-silico as a 12 cm long built in multi-lumen PC tubing (Young's modulus = 2 GPa) with the same design presented in Sect. \ref{Section: Fibre drawing}, but with varying depths, ranging from 0mm to 0.6mm. During the simulation, a range of loads were applied to its end, while the end deflection was measured. Displacement - Force relationships from this simulation are presented in Fig. \ref{fig:FEA1}b), with the bending stiffness computed using Eq. \ref{Eq:Bending stiffness}, and later plotted in Fig. \ref{fig:FEA1}c). From these plots, we observe a clear downward sloping relationship between the profile depth and bending stiffness. Thus, it can be concluded that having deeper grooves would lead to improved flexibility. While such an outcome is desirable, striking a balance between dexterity and support for the delivery of other endovascular medical devices is equally important. Therefore, we have opted to use 0.4mm depth for the second segment, as it must maneuver into difficult to reach parts of the body, while using a 0.2mm depth for the first segment, since the requirement for tight steering is less justified, while it may need to support heavier devices such as a stent graft or a stiff guidewire.

\subsection{Bending characterization}

A substantial advantage of a 2-segment steerable catheter is its ability to achieve multi-plane bending, while a standard steerable catheter can only bend in one direction. This enables the catheter to traverse challenging vessel geometries.

\begin{figure}
    \centering
    \includegraphics[width=0.46\textwidth]{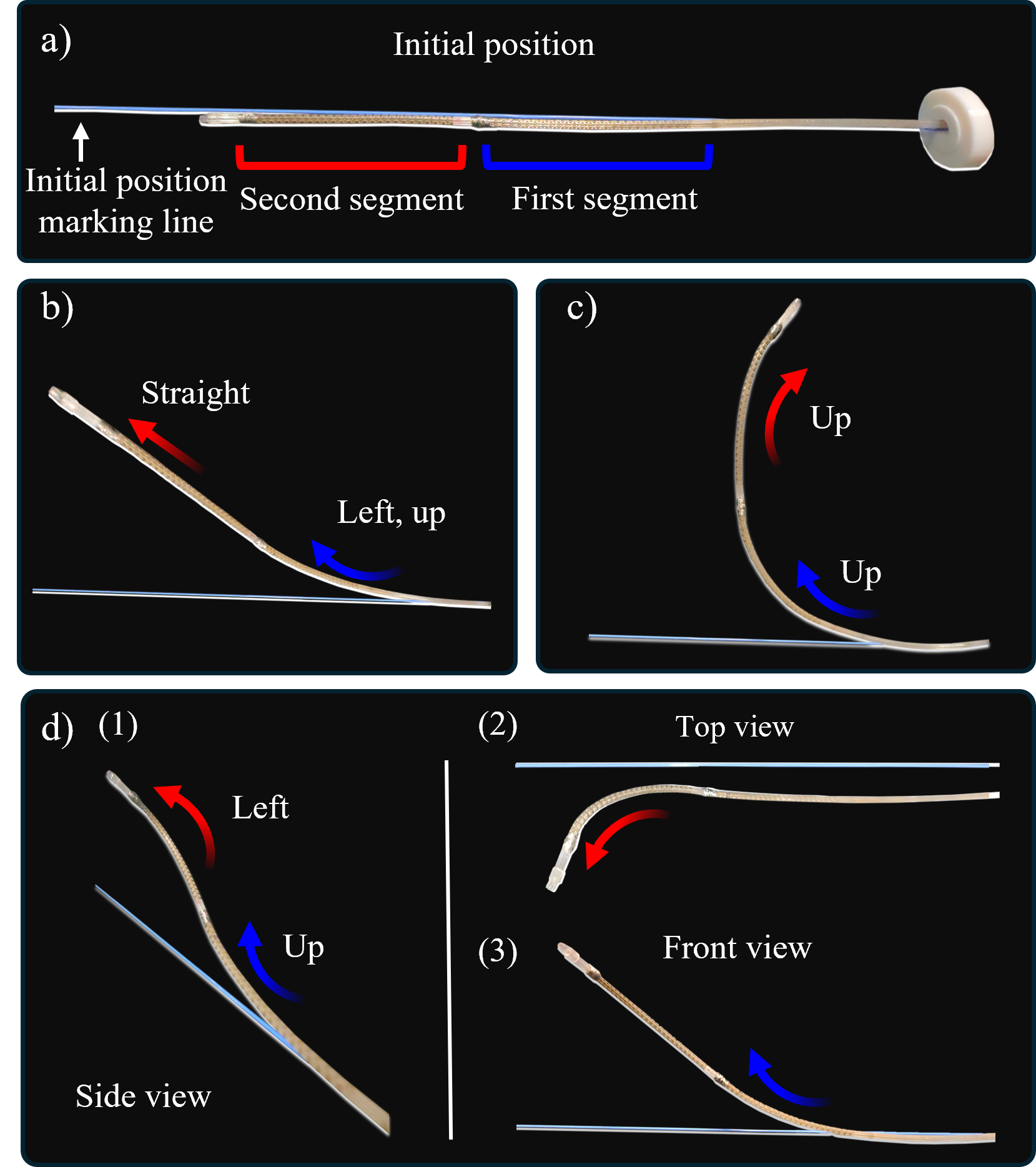}
    \caption{Demonstration of two-segment tendon-driven catheter bending. a) Initial pose of the robotic catheter. The blue line indicates the reference initial position. b) The second segment is steered toward the upper-left direction. c) Both the first and second segments are steered upward. d) The first segment is steered to the left, and the second segment is steered upward. Diagonal (1), top (2), and side (3) views are shown. }
    \label{fig:Bending}
\end{figure}

To illustrate the steerability of the tendon-driven catheter, we secured the handle on an optical table (Thorlabs, New Jersey, US), with the initial position aligned with a blue line (Fig.\ref{fig:Bending}a)). We then actuated the second catheter segment until it reaches its maximum position (Fig.\ref{fig:Bending}b)), with the tip reaching a maximum bending angle of $35^{\circ}$. Afterwards, we actuated the first and second segments jointly (Fig.\ref{fig:Bending}c)) reaching a maximum bending angle of $72^{\circ}$. To show multi-plane bending, we steer the second-segment to bend towards up side (up to $36^{\circ}$ with respect to the first segment), and the first segment towards left side (up to $38^{\circ}$). The diagonal, top, and side views are presented in Figure 5(d). 

\subsection{Navigation Characterization}

The key characteristic of a multi-segment steerable catheter is its ability to navigate complex and tortuous environments, such as those found in the human vasculature. 

To create an environment to validate our system's performance, we took inspiration from Kim et al.'s ring-based navigation course, used to test magnetic micro-catheters \cite{kim2019ferromagnetic}. 3D printed PLA rings were placed loosely across the table, creating a path for the catheter to traverse.

To demonstrate steering, we designed a ring obstacle course consisting of a C-shaped curve where the catheter must bend out of plane and bend downwards in order to pass through the rings. The distance between each ring is 10 cm, with the middle ring being 3 cm shorter than its two adjacent rings, which are 15 cm tall. The middle ring is moved 2.5 cm from the centerline. Rings are fastened to an optical table, while the robotic handle attached to a linear stage with a screw clamp mechanism, then it is advanced slowly using a linear stage, and steered using an XBox controller.

\begin{figure}
    \centering
    \includegraphics[width=0.46\textwidth]{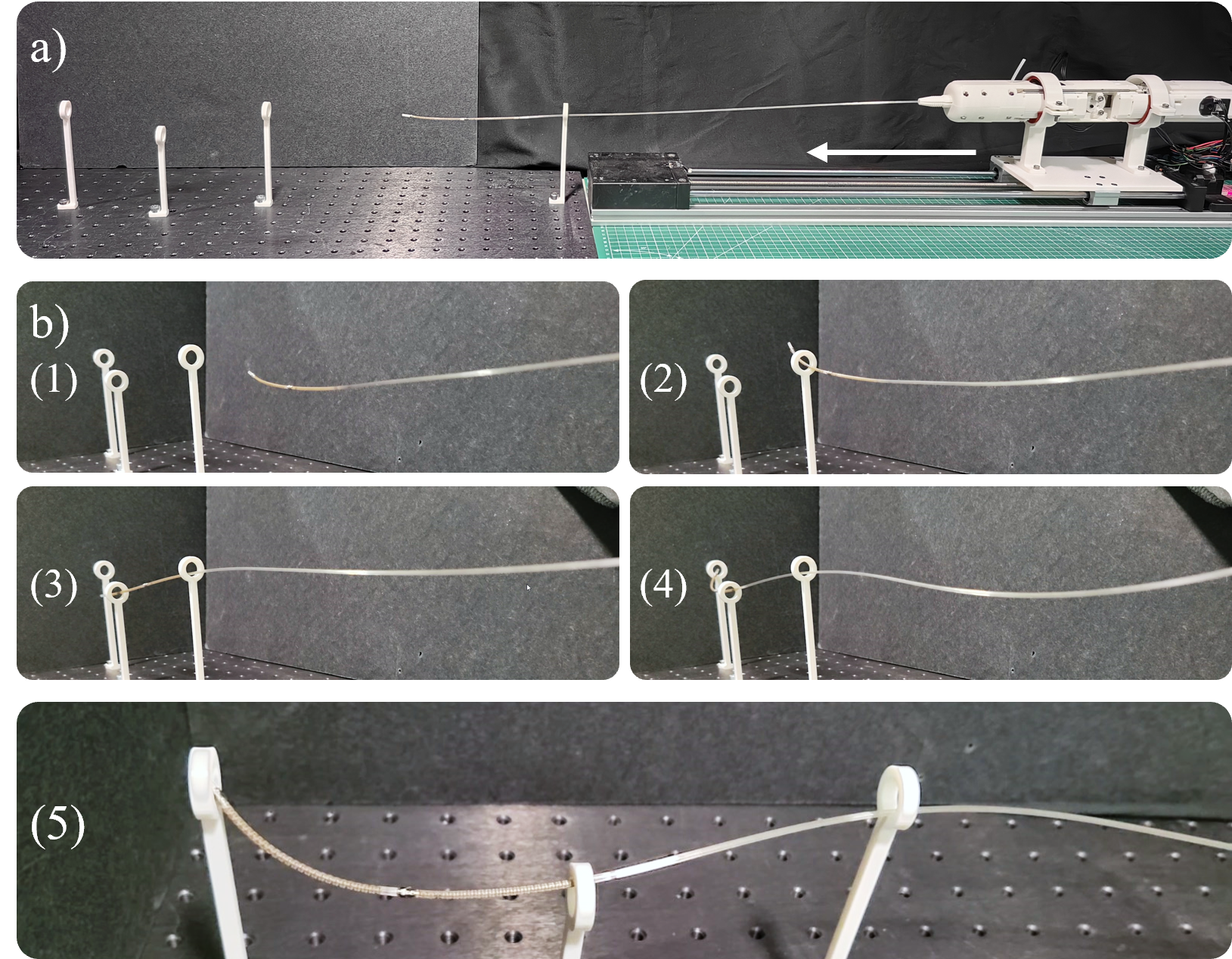}
    \caption{Experimental demonstration of tendon-driven catheter navigation. a) Experimental setup: The handle was fixed on a linear stage to enable catheter advancement. b) A series of images shows the robotic catheter navigating through the three target rings in sequence.}
    \label{fig:In-vitro}
\end{figure}

\begin{figure*}
    \centering
    \includegraphics[width=0.92\textwidth]{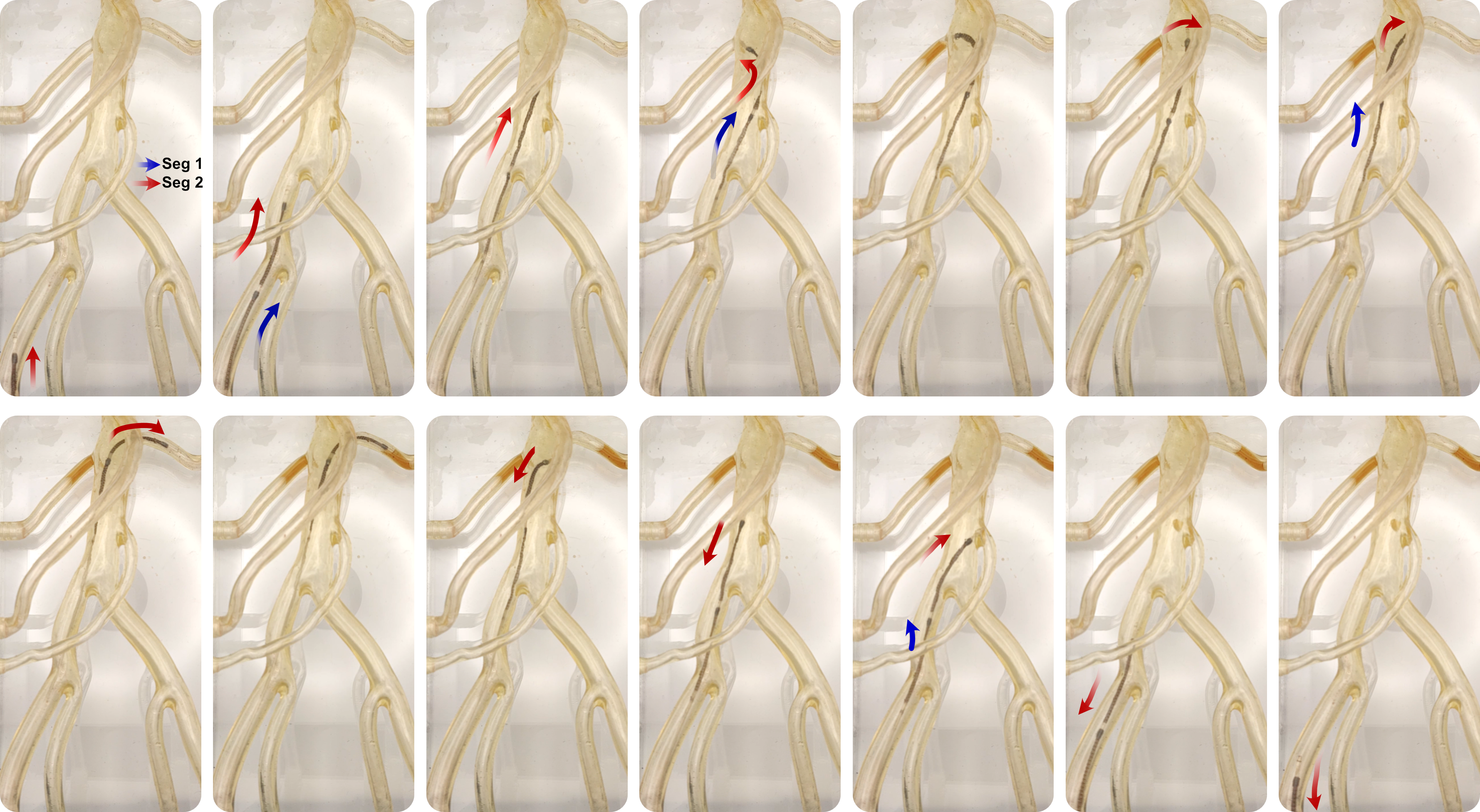}
    \caption{In-vitro evaluation of tendon driven catheter: Tendon-driven catheter navigating through an AAA phantom and delivering fluid (brown color) to different arterial branches. Blue and red arrows show the bending direction of the first (proximal) and second (distal) segment, respectively. }
    \label{fig:In-vitro}
\end{figure*}

\subsection{In-Vitro Study}

Finally, we extend our experiment to investigate the catheter's performance under a more realistic preclinical setting (Fig. \ref{fig:In-vitro}). A silicon abdominal aorta phantom (Elastrat, Geneva, Switzerland), consisting of the common iliac artery and the lower portion of the abdominal aorta, up until the celiac trunk, was used. While still a simplified model for the purposes of surgical training, this phantom replicates the complex anatomical environment of a true surgery. 

In this study, we would like to mimic a small portion of the FEVAR/BEVAR intervention, where we assume that the main stent graft is already in place, while we attempt to navigate the catheter to both renal arteries and the inferior mesenteric artery in order to deploy fenestrations/branches. Similar to the ring navigation experiment, the catheter is fastened to a linear stage, and remotely controlled using an Xbox controller, which reduces variations in user skill. 

To complete the procedure, colored dye was injected into the three arteries (similar to a thrombolysis procedure). The outcomes of this test can be visualized on a digital camera showing contrast agents in each indicated artery. 



\section{DISCUSSIONS}
In this work, we developed a 2-segment tendon driven catheter for AAA disease treatment. Our catheter fabrication technique enables low-cost manufacturing and mass production. The thermal drawing process allows us to fabricate tens of meters of catheter tubing in a single run. The robotic handle design follows a modular approach: four identical handle segments are assembled to achieve 4 active degrees of freedom for catheter control. This modular design simplifies assembly, and provides expandability, allowing it to accommodate for any number of even-numbered, antagonistically paired tendon-driven systems. The miniaturized handle is designed to be 3D printed in a single batch on one print plate, and its compact form factor allows surgeons to easily hold the entire handle in one hand.

The catheter features two serially connected bending segments, each providing two degrees of freedom. Compared to commercially available catheters, our catheter consists of an additional steerable segment, which improves its dexterity significantly, thus enhancing its accessibility. Following an initial FEM simulation, we recognized that by varying the profile depth, we can effectively reduce the bending stiffness by up to 83\%. However, upon recognizing that the catheter must also be able to provide support for other endovascular instruments, we opted for the 0.4 mm depth design in our final prototype at the distal segment, and 0.2 mm for the proximal segment. A limitation of the FEM study is the negligence of friction, which is an important topic that needs to be understood better in order for the catheter to be deployed into a clinical setting. Future works will revolve around the modeling of such forces and designing advanced feedforward models to compensate for their effects. 

During the bending experiment, we investigated the maximum angle of bending under benchtop situations, achieving a maximum bending angle of 72$^{\circ}$. From literature, average angles for the renal arteries, the superior mesenteric artery, as well as the iliac arteries are 64.1$^{\circ}$ \cite{stojadinovic2022anatomy}, 51.5$^{\circ}$ \cite{ozkurt2007sma}, 40.5$^{\circ}$ \cite{deswal2014study}, respectively. In general, the developed prototype is fit to traverse all of these vessels. This observation is confirmed in the in-vitro abdominal aorta phantom experiment. 

While the bending, navigation, and in vitro studies demonstrate the system's capabilities in navigating complex environments, several limitations must be highlighted. Most importantly, the catheter exhibits the muscling phenomenon, whereby the entire catheter shaft moves due to the pulling of the wires, and the curve alignment phenonemon, which refers to the undesired rotational motion of the full catheter \cite{US9844412B2}. In the previous works of Bogusky et al. \cite{US9844412B2} and Abdelaziz et al. \cite{abdelaziz2024thermally}, authors have tackled this problem using helical tendon channels, which not only redistributes the bending moments applied on the catheter along the longitudinal axis, thus counterbalances opposing compressive and tensile forces.

\section{CONCLUSION AND FUTURE WORK}

In this work, we present a 2 segment tendon-driven steerable catheter for navigating tortuous vasculature. The device consists of a thermally drawn multi-lumen tubing as its body, which was later laser profiled to reduce its bending stiffness. Then attached to a modular, expandable, and affordable actuation handle, fabricated using 3D printing. We evaluated the system's behavior first via an FEM simulation, ascertaining the impact of the profile, and deciding the appropriate depth to use. Thereafter, two experimental studies, designed to investigate the catheter's bending behavior, were carried out. The catheter underwent progressively more challenging tasks, traversing gentle bends in a ring obstacle course, then inside a realistic silicon phantom, cannulating into major arterial branches that must be protected during a BEVAR/FEVAR procedure. The proposed robotic catheter ideates a new class of handheld, highly dexterous instruments that simplified steering for clinicians, and has the potential to positively enhance precision and surgical outcomes.

Our future work will primarily focus on enhancing catheter design, advancing motion control, and integrating additional functionalities. Design optimisation will involve exploring softer materials, such as elastomers, for fabrication. Developing a helical-structured catheter in the passive section to further reduce undesired bending caused by tip actuation. Motion control will be improved by enabling more autonomous navigation, thereby reducing contact forces associated with manual operation. Additionally, we aim to integrate new functionalities, such as shape and proximity sensing, and integrate the catheter with other surgical devices.




 






\bibliographystyle{unsrt}
\bibliography{refs}
\vfill
\newpage

\vfill

\end{document}